\documentclass[runningheads]{llncs}

\usepackage{accv}

\usepackage{accvabbrv}
\usepackage{graphicx}
\usepackage{booktabs}
\usepackage{multirow}
\usepackage{amsmath}
\usepackage{amssymb}
\usepackage[accsupp]{axessibility}
\usepackage[breaklinks,colorlinks,citecolor=accvblue]{hyperref}
\usepackage{orcidlink}

\newcommand{\method}{MSPO}

\begin{document}

\title{Multimodal Semantic-Probabilistic Objectness for Open World Object Detection}
\titlerunning{\mbox{}}

\author{Weijun Tian, Rui Liu$^{*}$}
\authorrunning{\mbox{}}
\institute{School of Computer Science and Engineering, Beihang University\\
\email{\{tianweijun,lr\}@buaa.edu.cn}\\
$^{*}$Corresponding author.}

\maketitle
\pagestyle{plain}

\begin{abstract}
Open-world object detection (OWOD) requires a detector to recognize known categories, discover unnamed objects from unseen categories, and incrementally learn newly annotated classes.
PROB improves unknown discovery by modeling class-agnostic probabilistic objectness in the decoder-query space.
However, visual objectness alone cannot determine whether an object-like query corresponds to a hard known instance, an unseen-category object, or background clutter, resulting in an ambiguous known--unknown decision boundary.
We propose MSPO, a lightweight semantic calibration framework that augments PROB with task-aware known-category language priors while preserving its detector architecture and incremental learning protocol.
For each currently known category, MSPO constructs an extended text description covering category attributes, visual appearance, typical scenes, and functional usage, and encodes it using a frozen CLIP text encoder.
Decoder query features are projected into the same semantic space to estimate their support from the current known-category semantics.
This semantic evidence is fused with PROB's visual objectness to calibrate known and unknown predictions without turning OWOD into open-vocabulary classification.
Importantly, MSPO never uses future-category names, and all unseen categories remain unnamed during evaluation.
Experiments on M-OWODB and S-OWODB show that MSPO improves the strong PROB baseline on the main aggregate metrics while retaining competitive unknown recall.
It also improves early unknown-confusion metrics and raises PASCAL VOC final mAP by up to 2.7 points.
These results demonstrate that known-category language semantics provide an effective calibration signal for probabilistic objectness under the standard OWOD setting.
\end{abstract}

\section{Introduction}
\label{sec:introduction}

\begin{figure}[t]
\centering
\includegraphics[width=0.7\linewidth]{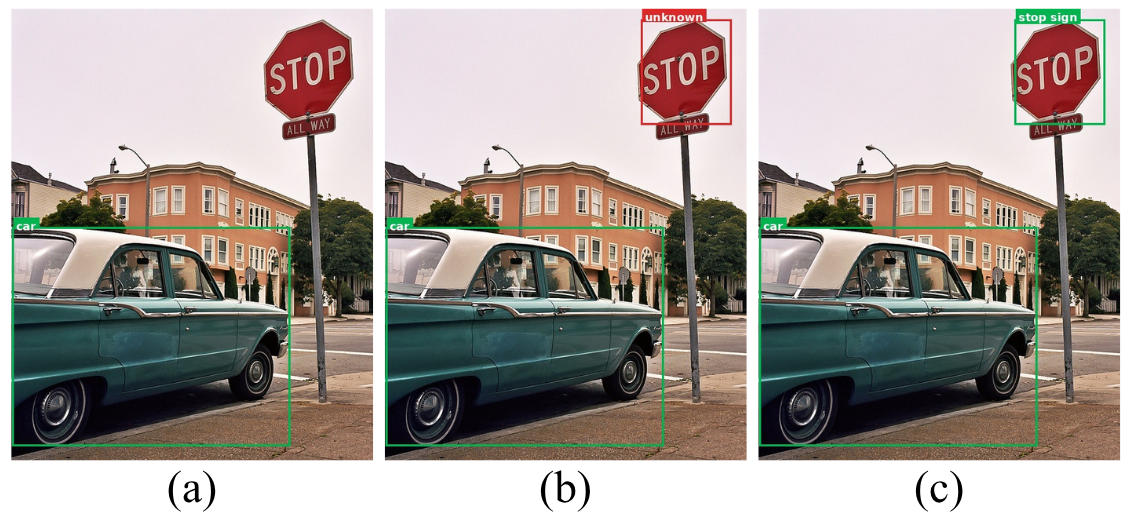}
\caption{Motivation of open world object detection. (a) A closed-set detector trained only with the known class car detects the car but ignores the unseen stop sign. (b) Before the stop sign category is introduced, an OWOD detector should localize it as an unnamed unknown object. (c) After incremental training with the newly introduced stop sign category, the same visual concept becomes a recognized known class.}
\label{fig:intro_case}
\end{figure}

Modern object detectors are usually trained under a closed-set assumption: all categories that may appear during inference are annotated during training.
This assumption is difficult to satisfy in open environments such as autonomous driving, robotics, surveillance, and embodied perception, where the visual world is not fixed and new categories may appear after deployment.
Open world object detection (OWOD) addresses this gap by requiring a detector to perform three abilities jointly: detect known objects, identify unknown objects that are not yet annotated, and incrementally learn newly introduced classes over time \cite{joseph2021ore,gupta2022owdetr,zohar2022prob}.
Figure~\ref{fig:intro_case} illustrates this protocol with a simple road-scene example.
A closed-set detector trained only on cars can localize the known car but ignores the unseen stop sign.
An OWOD detector should instead preserve the stop sign as an unnamed unknown object before the class is introduced, and should recognize it as a named class after incremental training.
This example highlights that OWOD is not merely a larger-vocabulary recognition problem, but a continual detection problem in which the detector must keep object-like evidence for future categories.

The main difficulty of OWOD lies in the ambiguous boundary among known objects, unknown objects, and background.
During training, categories outside the current known set are not annotated as unknown; consequently, a detector may treat hidden unknown objects as background.
ORE first formalizes OWOD and uses contrastive clustering, energy-based unknown identification, and exemplar replay to support open-world learning \cite{joseph2021ore}.
OW-DETR introduces a query-based transformer framework, using attention-driven pseudo-labeling and objectness scoring to mine potential unknown objects \cite{gupta2022owdetr}.
PROB further argues that pseudo-unknown labels are noisy and instead models probabilistic objectness in the decoder query embedding space \cite{zohar2022prob}.
This progression shows a clear trend: robust OWOD depends less on naming unknown categories and more on estimating whether a query corresponds to a valid object beyond the current label set.

Despite this progress, existing objectness-based OWOD methods remain primarily visual and statistical.
PROB estimates whether a query embedding belongs to the learned object distribution, but it does not explicitly ask whether the query is semantically supported by the known category space.
This distinction matters in open-world scenes.
A novel object can be visually salient and object-like while being semantically far from all known categories; conversely, a hard instance of a known class can have low classification confidence while still being close to its category semantics.
Without semantic evidence, a detector may absorb unknown objects into visually similar known classes or suppress them as background.

Vision-language models such as CLIP provide a complementary source of semantic structure by aligning visual concepts with natural language descriptions \cite{radford2021clip}.
Open-vocabulary detectors exploit this property to recognize classes specified by text at inference time \cite{gu2021vild,zhou2022regionclip,zareian2021ovrcnn,zhou2022detic,minderer2022simple,li2022glip,yao2023detclipv2,liu2023groundingdino}.
However, open-vocabulary detection and OWOD are not identical.
Open-vocabulary detection typically assumes that target class names are given, whereas OWOD must discover unknown objects before their names are available.
Therefore, rather than replacing an OWOD detector with a full open-vocabulary detector, we use language semantics as a calibration signal for objectness and the known versus unknown decision. Unlike open-vocabulary detectors, MSPO never scores a query against future or evaluation-only category names; text only forms semantic anchors for currently known categories, and unknown objects are still reported with a single unnamed unknown label.

\method{} constructs Extended-Text Embeddings for known categories using a frozen CLIP text encoder, projects transformer decoder queries into the same semantic space, and estimates the degree to which each query is supported by the current known-category semantics.
The complement of this support is converted into a semantic rejection term and fused with probabilistic objectness inside the trainable SPOF objectness loss.
The design keeps the PROB detector intact, adds only a small query-level semantic head, and allows each component to be independently enabled for ablation.

The central idea is simple: probabilistic objectness estimates whether a query is visually object-like, while known-semantic support estimates whether the query is explained by the current known categories.
Their combination better matches the OWOD objective, because unknown objects should be preserved when they are object-like but unsupported by the current known semantic space.
This formulation also gives a clear experimental question: under the same PROB-aligned OWOD and incremental detection protocols, can frozen language semantics improve the known--unknown trade-off without replacing the detector architecture?

Our contributions are summarized in three points:
\begin{itemize}
    \item We add Semantic Projection and Query-Text Alignment as a lightweight semantic branch for the classification head. By aligning decoder queries with current known-class text prototypes, it calibrates class logits and improves known-class recognition without changing the detector architecture.
    \item We propose Semantic-Probabilistic Objectness Fusion, which combines visual objectness with known-semantic support to distinguish known objects, potential unknown objects, and low-objectness background. This design gives the detector an unknown-aware objectness cue while keeping all future categories unnamed under the OWOD protocol.
    \item We evaluate \method{} under the PROB-aligned protocols on M-OWODB, S-OWODB, and PASCAL VOC incremental detection. Across these benchmarks, \method{} consistently improves the strong PROB baseline, and the ablations verify the contributions of semantic alignment and objectness fusion.
\end{itemize}

\section{Related Work}
\label{sec:related_work}

\subsection{Open World Object Detection}
Open world object detection extends closed-set detection by introducing unknown discovery and incremental learning into the detection protocol \cite{joseph2021ore}.
Modern OWOD systems build on two-stage, dense, and transformer detectors \cite{girshick2014rcnn,girshick2015fastrcnn,ren2015fasterrcnn,liu2016ssd,redmon2016yolo,lin2017focalloss,tian2019fcos,he2017maskrcnn,carion2020detr,zhu2021deformable,zhang2022dino}.
ORE \cite{joseph2021ore} defines the task with energy-based unknown identification, contrastive clustering, and exemplar replay; OW-DETR \cite{gupta2022owdetr} mines pseudo unknown objects with transformer attention; PROB \cite{zohar2022prob} avoids noisy pseudo-unknown supervision by learning probabilistic objectness from query embeddings.
Our work follows the PROB evaluation lineage but adds language-guided semantic evidence to query-level objectness estimation.

\subsection{Objectness, Open-Set Recognition, and Unknown Discovery}
Objectness distinguishes foreground objects from background, but in OWOD it must also preserve objects outside the current annotation set.
Energy-based detection, pseudo-label mining, and probabilistic query modeling provide visual routes to this goal \cite{joseph2021ore,gupta2022owdetr,zohar2022prob}.
Open-set and out-of-distribution recognition study rejection with activations, confidence calibration, or energy scores \cite{bendale2016openmax,hendrycks2017baseline,liang2018odin,liu2020energyood}, but OWOD also requires localization, background separation, and later class learning.
We therefore use known-semantic support as a complementary query-level calibration signal rather than a replacement for objectness.

\subsection{Vision-Language and Open-Vocabulary Detection}
Vision-language pretraining, especially CLIP \cite{radford2021clip}, provides transferable image-text representations.
Open-vocabulary detectors use such representations through distillation, region-language alignment, caption supervision, or grounded pretraining \cite{gu2021vild,zhou2022regionclip,zareian2021ovrcnn,zhou2022detic,yao2023detclipv2,liu2023groundingdino,kirillov2023sam,oquab2023dinov2,caron2021dino}.
These methods show the value of language semantics, but often change the detector family or assume external vocabulary supervision.
In contrast, \method{} uses frozen Extended-Text Embeddings as semantic anchors inside a PROB-style OWOD detector, keeping the comparison focused on probabilistic objectness.

\subsection{Incremental Detection and Representation Learning}
OWOD also requires incremental learning: unknown categories in one task may become known later.
This connects to distillation, exemplar replay, representation regularization, and contrastive representation learning \cite{hinton2015distill,li2018lwf,rebuffi2017icarl,shmelkov2017ilod,he2020moco,chen2020simclr,khosla2020supcon}.
Our method keeps the incremental protocol unchanged and supplies a semantic reference space updated by the active known-class set.

\section{Method}
\label{sec:method}

\subsection{Problem Setting and Design Principle}
We follow the standard OWOD protocol introduced by ORE and used by OW-DETR and PROB.
At task $t$, the detector is trained with annotations for the currently known class set $\mathcal{C}^{seen}_t$.
Objects from classes outside $\mathcal{C}^{seen}_t$ may appear during testing, but their names are not provided to the detector and they should be reported as \emph{unknown}.
After each task, a subset of unknown categories becomes annotated and is added to the known set for the next task.
The goal is therefore not open-vocabulary naming, but a joint optimization of known-class detection, unknown-object discovery, and incremental learning.

Our design starts from the probabilistic view of PROB rather than replacing it.
Given an image, a query-based detector produces $N$ decoder query features $\{q_i\}_{i=1}^{N}$.
Each query predicts a bounding box, known-class logits, and an objectness score.
PROB decomposes the probability of assigning label $l$ to query $q_i$ as
\begin{equation}
    p(l \mid q_i) = p(l \mid o_i, q_i)\,p(o_i \mid q_i),
    \label{eq:prob_factorization}
\end{equation}
where $o_i$ denotes the event that the query corresponds to a foreground object.
This factorization is well suited for OWOD because an unknown instance has no known class label, but it should still receive high objectness.
The limitation is that the objectness term is estimated from visual query statistics alone.
\method{} keeps this factorization and asks a more specific question: can known-class language semantics provide a complementary query-level cue for calibrating $p(o_i\mid q_i)$ without naming future-category classes?

\subsection{Overview of MSPO}
\label{sec:overview}

\begin{figure}[t]
\centering
\includegraphics[width=\linewidth]{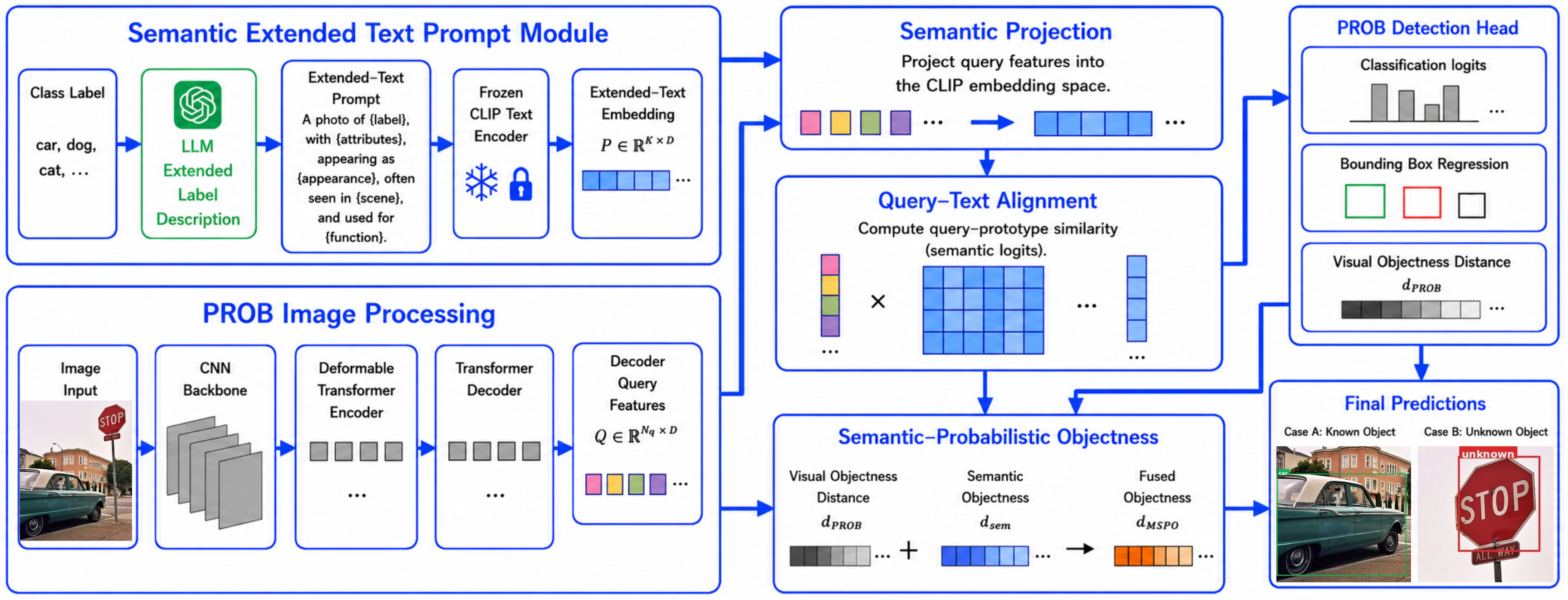}
\caption{Overview of \method{}. Extended text prompts are encoded by a frozen CLIP text encoder into $P\in\mathbb{R}^{K\times D_t}$, while PROB extracts decoder query features $Q\in\mathbb{R}^{N\times D_q}$. Semantic Projection and Query-Text Alignment compute known-semantic support, and SPOF fuses visual objectness distance with semantic rejection to form $D_{\mathrm{SPOF}}$ and $\mathcal{L}_{obj}$. The PROB detection head keeps classification and box prediction under the shared objective.}
\label{fig:framework}
\end{figure}

As shown in Fig.~\ref{fig:framework}, \method{} follows the same information flow as PROB and adds a lightweight semantic branch.
The PROB Image Processing stream extracts Decoder Query Features $Q\in\mathbb{R}^{N\times D_q}$, where $N$ is the number of decoder queries and $D_q$ is the detector-query feature dimension.
These queries are fed to the PROB Detection Head for classification logits, bounding box regression, and visual objectness distance.
The semantic stream has three roles.
First, the Semantic Extended Text Prompt constructs known-class semantic anchors as an Extended-Text Embedding matrix $P\in\mathbb{R}^{K\times D_t}$, where $K=|\mathcal{C}^{seen}_t|$ and $D_t$ is the CLIP text embedding dimension.
Second, Semantic Projection maps query features from $D_q$ to $D_t$, and Query-Text Alignment computes query-prototype semantic logits.
Third, Semantic-Probabilistic Objectness Fusion converts weak known-class semantic support into a semantic rejection term and fuses it with the visual objectness distance in the trainable objectness loss.
The branch is task-aware: at task $t$, it uses only $\mathcal{C}^{seen}_t$.
It does not introduce future-category names and therefore remains an OWOD method rather than an open-vocabulary detector.

\subsection{Semantic Extended Text Prompt}
\label{sec:extended_prompt}
For each known class $c \in \mathcal{C}^{seen}_t$, we construct an Extended-Text Prompt $r_c$ from its class name and an offline generated extended description.
The prompt follows the fixed pattern:
\begin{quote}
\small
``a photo of \{label\}, with \{attributes\}, appearing as \{appearance\}, often seen in \{scene\}, and used for \{function\}.''
\end{quote}
The fields describe category attributes, typical appearance, common scene context, and functional usage.
For reproducibility, the descriptions are generated offline before training, fixed once a category becomes known, and reused in later tasks.
At task $t$, prompts are constructed only for classes in $\mathcal{C}^{seen}_t$; future-category names are never queried, encoded, or used as text prototypes.
The complete prompt list can be provided in the supplementary material.
A frozen CLIP text encoder $E_{txt}$ maps each prompt to a normalized Extended-Text Embedding:
\begin{equation}
    p_c = \frac{E_{txt}(r_c)}{\|E_{txt}(r_c)\|_2}, \qquad c \in \mathcal{C}^{seen}_t.
    \label{eq:text_embedding}
\end{equation}
Stacking all $K$ embeddings gives $P=[p_1,\ldots,p_K]^\top\in\mathbb{R}^{K\times D_t}$.
Because the CLIP text encoder is frozen, $P$ is pre-computed at model construction and reused during training and evaluation.
This module provides semantic anchors for the current known categories, not an external vocabulary for recognizing future-category objects.

\subsection{Semantic Projection and Query-Text Alignment}
The semantic branch operates on the same Decoder Query Features used by the detector.
Let $Q=[q_1,\ldots,q_N]^\top\in\mathbb{R}^{N\times D_q}$ denote the query matrix.
Semantic Projection uses a lightweight projection head $g:\mathbb{R}^{D_q}\rightarrow\mathbb{R}^{D_t}$ to map each query into the CLIP text embedding space:
\begin{equation}
    h_i = \frac{g(q_i)}{\|g(q_i)\|_2}, \qquad h_i\in\mathbb{R}^{D_t}.
    \label{eq:query_projection}
\end{equation}
In our implementation, $g$ is a two-layer MLP, Linear--ReLU--Dropout--Linear, with dropout $p_{drop}=0.1$.
It maps the detector hidden dimension to the 512-dimensional CLIP text space.
The CLIP text encoder is frozen; gradients from the semantic loss update the detector and the projection head, but not the text encoder.
Query-Text Alignment computes semantic logits between projected queries and Extended-Text Embeddings:
\begin{equation}
    a_{i,c} = \frac{h_i^{\top}p_c}{\gamma_s}, \qquad c \in \mathcal{C}^{seen}_t,
    \label{eq:semantic_logits}
\end{equation}
where $\gamma_s$ is the semantic temperature.
For matched object queries, the alignment head is supervised by the current ground-truth class:
\begin{equation}
    \mathcal{L}_{sem} = \frac{1}{|\mathcal{M}|}\sum_{(i,y_i)\in\mathcal{M}} \mathrm{CE}(a_i,y_i),
    \label{eq:semantic_loss}
\end{equation}
where $\mathcal{M}$ is the set of Hungarian-matched object queries.
We intentionally do not assign a semantic background label to unmatched queries, because unmatched queries may contain hidden unknown objects under the OWOD protocol.
This loss makes the query representation aware of known-class semantic structure while preserving the detector's original supervision.
It is not applied to future-category classes because their names are unavailable during the corresponding OWOD task.

\subsection{Semantic-Probabilistic Objectness Fusion}
\label{sec:semantic_prob_objectness}
The goal of SPOF is to give the detector a query-level criterion for separating three cases: known objects, unknown objects, and background.
Visual objectness alone can separate many objects from background, but it cannot decide whether an object-like query belongs to the known semantic space.
Conversely, weak known-class semantic support alone is not sufficient evidence for an unknown object, because background regions may also be semantically unsupported.
SPOF therefore combines the two cues: a query is a plausible unknown only when it is visually object-like and weakly supported by all current known classes.

The PROB Detection Head first measures whether a query is object-like by its distance to a class-agnostic object-query distribution.
Let $\mu \in \mathbb{R}^{D_q}$ and $\Sigma \in \mathbb{R}^{D_q\times D_q}$ be the running mean and covariance of object query features.
The visual objectness distance is
\begin{equation}
    d_i^{\mathrm{PROB}} = (q_i - \mu)^{\top}\Sigma^{-1}(q_i - \mu),
    \label{eq:prob_distance}
\end{equation}
and its bounded visual objectness evidence is
\begin{equation}
    V_i = \exp(-\tau d_i^{\mathrm{PROB}}),
    \label{eq:visual_objectness}
\end{equation}
where $\tau$ is the objectness temperature.
A low $d_i^{\mathrm{PROB}}$ and high $V_i$ indicate an object-like query, while background clutter usually has a large distance and low visual evidence.

To distinguish known objects from potential unknown objects, we estimate known-semantic support from the query-text logits:
\begin{equation}
    K_i^{sem} = \rho_s\max_{c\in\mathcal{C}^{seen}_t}\mathrm{softmax}(a_i)_c
    + (1-\rho_s)\sigma\left(\max_{c\in\mathcal{C}^{seen}_t}a_{i,c}\right).
    \label{eq:known_support}
\end{equation}
Here $\rho_s\in[0,1]$ balances relative support within the current known set and absolute query-text affinity; we set $\rho_s=0.5$.
High $K_i^{sem}$ means that the query is well explained by at least one current known class.
Its complement is the known-semantic rejection term
\begin{equation}
    R_i^{sem} = 1-K_i^{sem},
    \label{eq:semantic_rejection}
\end{equation}
which becomes large when the query is poorly supported by the known semantic anchors.
SPOF uses the product of visual objectness and semantic rejection as the unknown-aware objectness cue:
\begin{equation}
    U_i^{\mathrm{SPOF}} = V_i R_i^{sem}.
    \label{eq:spof_unknownness}
\end{equation}
This product encodes the desired OWOD boundary: known objects have high $V_i$ but low $R_i^{sem}$, unknown objects should have high $V_i$ and high $R_i^{sem}$, and background regions are suppressed by low $V_i$ even if their semantic rejection is high.

During training, future unknown labels are unavailable, so SPOF supervises the learnable boundary using matched known objects.
The fused distance for objectness learning is
\begin{equation}
    D_i^{\mathrm{SPOF}} = d_i^{\mathrm{PROB}} + \lambda_s R_i^{sem},
    \label{eq:spof_distance}
\end{equation}
where $\lambda_s$ controls how strongly weak known-semantic support is penalized.
The SPOF objectness loss averages this fused distance over Hungarian-matched object queries:
\begin{equation}
    \mathcal{L}_{obj} = \frac{1}{|\mathcal{M}|}\sum_{i\in\mathcal{M}}D_i^{\mathrm{SPOF}},
    \label{eq:mspo_obj_loss}
\end{equation}
so minimizing $\mathcal{L}_{obj}$ simultaneously pulls known objects toward the visual object distribution and increases their known-semantic support.
After such training, high objectness with low rejection is treated as known-object evidence, high objectness with high rejection becomes unknown-object evidence, and low objectness remains background.
The full training objective uses the semantic alignment loss before the SPOF objectness loss:
\begin{equation}
    \mathcal{L} = \mathcal{L}_{cls} + \lambda_1\mathcal{L}_{box}
    + \lambda_2\mathcal{L}_{sem} + \lambda_3\mathcal{L}_{obj}.
    \label{eq:total_loss}
\end{equation}
Here $\mathcal{L}_{box}$ denotes the bounding-box regression loss used by the detector, while $\lambda_1$, $\lambda_2$, and $\lambda_3$ balance localization, semantic alignment, and SPOF objectness.
The detector can still use the semantic logits for classification calibration with weight $\alpha_s$, but the trainable SPOF module is fully specified by Eqs.~\eqref{eq:known_support}--\eqref{eq:total_loss}.
Setting $\lambda_2=0$, $\alpha_s=0$, and $\lambda_s=0$ reduces $\mathcal{L}_{obj}$ to the original PROB visual objectness objective.
All semantic modules are controlled by independent weights and switches, enabling ablations for label-only prompt, extended prompt, objectness-fusion-only, and full MSPO variants.
The method does not classify unknown objects into external text labels; all future-category objects remain unnamed unknowns during evaluation.

\section{Experiments}
\label{sec:experiments}

\subsection{Datasets and Protocols}
We follow the dataset splits and reporting protocol used by PROB, which are inherited from ORE and OW-DETR.
The main OWOD evaluation uses M-OWODB and S-OWODB.
M-OWODB is built from MS-COCO and PASCAL VOC \cite{lin2014coco,everingham2010pascal} and follows the superclass-mixed split, where the 80 object categories are divided into four incremental tasks with 20 new classes per task.
S-OWODB uses MS-COCO \cite{lin2014coco} with superclass-separated task splits, making semantic transfer across tasks more challenging.
At task $t$, classes from previous tasks are treated as previously known, classes introduced in the current task are currently known, and future-task classes are treated as unknown during evaluation.
The detector is evaluated on all introduced known classes and on its ability to retrieve unlabeled unknown objects.

We also evaluate incremental object detection on PASCAL VOC 2007 following PROB.
The 20 VOC classes are split into two-stage protocols: $10+10$, $15+5$, and $19+1$.
The first number denotes base classes learned in the first stage, and the second number denotes new classes introduced in the second stage.
This experiment tests whether semantic calibration preserves PROB's incremental learning behavior rather than only improving unknown discovery.

\subsection{Metrics and Implementation Details}
Following PROB, ORE, and OW-DETR, we report unknown recall at the top 50 detections (U-Recall), AP at IoU 0.5 for known classes, Wilderness Impact (WI), and Absolute Open-Set Error (A-OSE).
For Tasks 2--4, known AP is decomposed into previously known, currently known, and all known categories.
For PASCAL VOC incremental detection, we report AP for old classes, AP for new classes, and final mAP.

The detector is implemented with the Deformable DETR based PROB codebase, using the same task splits, evaluation code, and incremental training schedule as PROB.
We use CLIP ViT-B with patch size 16 to construct $D_t=512$ dimensional Extended-Text Embeddings and keep the text encoder frozen; CLIP is used only to pre-compute text embeddings at model construction.
The semantic projection head uses dropout $p_{drop}=0.1$ and the semantic temperature is $\gamma_s=0.07$.
For MSPO, we set the semantic logit weight $\alpha_s=0.10$, the semantic distance-fusion weight $\lambda_s=0.10$, the semantic alignment loss weight $\lambda_2=0.10$, the support mixture weight $\rho_s=0.50$, and the semantic unknown inference weight to $0.50$.
Following PROB, the fused objectness loss coefficient is $\lambda_3=8\times10^{-4}$ and the objectness temperature is $\tau=1.3$; the box loss weight $\lambda_1$ follows the default setting in the released PROB code.
We use the same optimizer, learning-rate schedule, fine-tuning splits, and evaluation metrics as the PROB aligned scripts, and all comparisons are made under this shared protocol.

\subsection{Main Results on M-OWODB and S-OWODB}
Table~\ref{tab:main_owodb} reports the main OWOD results on M-OWODB and S-OWODB in the same PROB format.
We compare against the same method lineage used by PROB, including ORE, OW-DETR, and PROB; for M-OWODB, we also include intermediate OWOD methods reported in the PROB comparison table.
\method{} improves the strongest aggregate metrics over PROB on both benchmarks, while a few current-class or unknown-recall entries remain close to PROB.
This matches our goal of strengthening a strong probabilistic OWOD baseline rather than replacing it with an open-vocabulary detector.

\begin{table}[t]
\centering
\caption{Main OWOD comparison under the PROB reporting protocol. The upper block reports M-OWODB and the lower block reports S-OWODB. AP and mAP are reported at IoU 0.5. Higher is better for all entries.}
\label{tab:main_owodb}
\resizebox{\linewidth}{!}{%
\begin{tabular}{lccccccccccccc}
\toprule
Method & \multicolumn{2}{c}{Task 1} & \multicolumn{4}{c}{Task 2} & \multicolumn{4}{c}{Task 3} & \multicolumn{3}{c}{Task 4} \\
\cmidrule(lr){2-3}\cmidrule(lr){4-7}\cmidrule(lr){8-11}\cmidrule(lr){12-14}
 & U-Recall & \multicolumn{1}{c}{mAP} & U-Recall & \multicolumn{3}{c}{mAP} & U-Recall & \multicolumn{3}{c}{mAP} & \multicolumn{3}{c}{mAP} \\
\cmidrule(lr){3-3}\cmidrule(lr){5-7}\cmidrule(lr){9-11}\cmidrule(lr){12-14}
 & & Cur. & & Pre. & Cur. & Both & & Pre. & Cur. & Both & Pre. & Cur. & Both \\
\midrule
\multicolumn{14}{l}{\textbf{M-OWODB}} \\
ORE & 4.9 & 56.0 & 2.9 & 52.7 & 26.0 & 39.4 & 3.9 & 38.2 & 12.7 & 29.7 & 29.6 & 12.4 & 25.3 \\
UC-OWOD & 2.4 & 50.7 & 3.4 & 33.1 & 30.5 & 31.8 & 8.7 & 28.8 & 16.3 & 24.6 & 25.6 & 15.9 & 23.2 \\
OCPL & 8.3 & 56.6 & 7.7 & 50.6 & 27.5 & 39.1 & 11.9 & 38.7 & 14.7 & 30.7 & 30.7 & 14.4 & 26.7 \\
2B-OCD & 12.1 & 56.4 & 9.4 & 51.6 & 25.3 & 38.5 & 11.6 & 37.2 & 13.2 & 29.2 & 30.0 & 13.3 & 25.8 \\
OW-DETR & 7.5 & 59.2 & 6.2 & 53.6 & 33.5 & 42.9 & 5.7 & 38.3 & 15.8 & 30.8 & 31.4 & 17.1 & 27.8 \\
PROB & 19.4 & 59.5 & 17.4 & 55.7 & 32.2 & 44.0 & 19.6 & 43.0 & 22.2 & 36.0 & 35.7 & 18.9 & 31.5 \\
\method{} & \textbf{20.1} & \textbf{61.7} & \textbf{17.8} & \textbf{57.5} & 32.0 & \textbf{45.4} & \textbf{20.2} & \textbf{44.5} & \textbf{22.8} & \textbf{37.0} & \textbf{36.9} & 18.7 & \textbf{32.3} \\
\midrule
\multicolumn{14}{l}{\textbf{S-OWODB}} \\
ORE & 1.5 & 61.4 & 3.9 & 56.5 & 26.1 & 40.6 & 3.6 & 38.7 & 23.7 & 33.7 & 33.6 & 26.3 & 31.8 \\
OW-DETR & 5.7 & 71.5 & 6.2 & 62.8 & 27.5 & 43.8 & 6.9 & 45.2 & 24.9 & 38.5 & 38.2 & 28.1 & 33.1 \\
PROB & 17.6 & 73.4 & 22.3 & 66.3 & 36.0 & 50.4 & 24.8 & 47.8 & 30.4 & 42.0 & 42.6 & 31.7 & 39.9 \\
\method{} & \textbf{18.2} & \textbf{76.1} & \textbf{22.9} & \textbf{68.4} & \textbf{36.6} & \textbf{51.9} & 24.6 & \textbf{49.1} & \textbf{31.0} & \textbf{43.1} & \textbf{43.8} & 31.5 & \textbf{40.8} \\
\bottomrule
\end{tabular}}
\end{table}

\paragraph{Analysis.}
The largest gains appear in early known-class recognition and in aggregate known-class AP.
On M-OWODB, \method{} improves Task 1 current AP by +2.2 points and Task 2 Both AP by +1.4 points over PROB; it also improves Task 4 previously known AP by +1.2 points.
On S-OWODB, the strongest gains are Task 1 current AP by +2.7 points, Task 2 previously known AP by +2.1 points, and Task 2 Both AP by +1.5 points.
Averaged over the reported entries, \method{} improves PROB by about +0.7 points on M-OWODB and +0.8 points on S-OWODB.
The few lower entries, such as M-OWODB Task 2 current AP and S-OWODB Task 3 U-Recall, indicate that semantic calibration mainly improves the overall known and unknown trade-off rather than uniformly increasing every sub-metric.

\subsection{PASCAL VOC Incremental Object Detection}
Table~\ref{tab:pascal_voc} reports the two-stage PASCAL VOC incremental detection results following PROB.
The comparison uses the same old-class, new-class, and final mAP format and includes OW-DETR and PROB.
\method{} improves final mAP over PROB in all three splits, indicating that semantic calibration does not compromise incremental learning.

\begin{table}[t]
\centering
\caption{PASCAL VOC 2007 two-stage incremental object detection. The setting $a+b$ means that $a$ base classes are learned first and $b$ new classes are introduced in the second stage.}
\label{tab:pascal_voc}
\begin{tabular}{llccc}
\toprule
Setting & Method & Old & New & Final mAP \\
\midrule
\multirow{3}{*}{$10+10$}
& OW-DETR & 63.5 & 67.9 & 65.7 \\
& PROB & 66.0 & 67.2 & 66.5 \\
& \method{} & \textbf{68.2} & \textbf{68.0} & \textbf{68.4} \\
\midrule
\multirow{3}{*}{$15+5$}
& OW-DETR & 72.2 & 59.8 & 69.4 \\
& PROB & 73.2 & 60.8 & 70.1 \\
& \method{} & \textbf{75.9} & 60.2 & \textbf{72.7} \\
\midrule
\multirow{3}{*}{$19+1$}
& OW-DETR & 70.2 & 62.0 & 70.2 \\
& PROB & 73.9 & 48.5 & 72.6 \\
& \method{} & \textbf{76.6} & 47.9 & \textbf{75.3} \\
\bottomrule
\end{tabular}
\end{table}

\paragraph{Analysis.}
The clearest Pascal VOC gains appear in the $15+5$ and $19+1$ splits, where final mAP improves by +2.6 and +2.7 points over PROB, respectively.
In the $19+1$ split, old-class AP also increases by +2.7 points, suggesting stronger retention of previously learned categories.
Across the three incremental settings, the average final mAP gain is about +2.4 points.
New-class AP is slightly lower than PROB in the $15+5$ and $19+1$ splits, indicating that the main benefit of MSPO comes from old-class stability and final detection quality rather than a uniform increase on every class group.

\subsection{Unknown Confusion Analysis}
Known-semantic calibration is intended to reduce the confusion between unknown objects and known classes.
Following PROB, we report WI and A-OSE on the first three M-OWODB tasks, where future-category objects exist.

\begin{table}[t]
\centering
\caption{Unknown confusion analysis on M-OWODB. Lower is better for WI and A-OSE, while higher is better for U-Recall.}
\label{tab:unknown_confusion}
\resizebox{\linewidth}{!}{%
\begin{tabular}{lccccccccc}
\toprule
Method & \multicolumn{3}{c}{Task 1} & \multicolumn{3}{c}{Task 2} & \multicolumn{3}{c}{Task 3} \\
\cmidrule(lr){2-4}\cmidrule(lr){5-7}\cmidrule(lr){8-10}
 & U-Recall & WI & A-OSE & U-Recall & WI & A-OSE & U-Recall & WI & A-OSE \\
\midrule
ORE & 4.9 & 0.062 & 10459 & 2.9 & 0.028 & 10445 & 3.9 & 0.021 & 7990 \\
OW-DETR & 7.5 & 0.057 & 10240 & 6.2 & 0.028 & 8441 & 5.7 & 0.016 & 6803 \\
PROB & 19.4 & 0.057 & 5195 & 17.4 & 0.034 & 6452 & 19.6 & \textbf{0.015} & \textbf{2641} \\
\method{} & \textbf{20.1} & \textbf{0.055} & \textbf{5010} & \textbf{17.8} & \textbf{0.033} & \textbf{6260} & \textbf{20.2} & \textbf{0.015} & 2685 \\
\bottomrule
\end{tabular}}
\end{table}

Table~\ref{tab:unknown_confusion} shows that \method{} improves U-Recall in all three tasks, with the largest gain of +0.7 points in Task 1.
WI is slightly reduced in Tasks 1 and 2 and remains tied with PROB in Task 3.
A-OSE decreases by 185 and 192 errors in the first two tasks, while Task 3 is slightly higher than PROB.
This trend is consistent with a calibration module: semantic support improves the early known and unknown boundary, but it does not eliminate all late-stage open-set confusion.

\subsection{Ablation Analysis}
The architecture of \method{} is modular, allowing each semantic component to be enabled independently.
Following the component-wise ablation style commonly used in open-vocabulary and open-world studies, Table~\ref{tab:ablation} reports both the enabled modules and the resulting M-OWODB performance.
The Label-only Prompt variant uses the plain CLIP text prompt ``a photo of \{label\}'' and therefore tests whether class-name text anchoring alone changes the detector.
SETP replaces this plain prompt with the Extended-Text Prompt, while SPOF adds Semantic-Probabilistic Objectness Fusion.

\begin{table}[t]
\centering
\caption{Ablation study on M-OWODB. The Label-only Prompt variant uses the plain CLIP prompt ``a photo of \{label\}''. SETP denotes Semantic Extended Text Prompt, and SPOF denotes Semantic-Probabilistic Objectness Fusion.}
\label{tab:ablation}
\resizebox{\linewidth}{!}{%
\begin{tabular}{lccc cccc ccc}
\toprule
Method & Label-only & SETP & SPOF & T1 U-R & T1 mAP & T2 U-R & T2 Both & T3 U-R & T3 Both & T4 Both \\
\midrule
PROB & -- & -- & -- & 19.4 & 59.5 & 17.4 & 44.0 & 19.6 & 36.0 & 31.5 \\
Label-only Prompt & \checkmark & -- & -- & 19.2 & 59.7 & 17.3 & 44.1 & 19.4 & 36.1 & 31.3 \\
SETP & \checkmark & \checkmark & -- & 19.8 & 60.7 & 17.6 & 44.8 & 19.8 & 36.5 & 31.9 \\
SPOF & \checkmark & -- & \checkmark & 20.0 & 60.4 & \textbf{18.0} & 45.0 & 20.0 & 36.8 & 32.0 \\
\method{} & \checkmark & \checkmark & \checkmark & \textbf{20.1} & \textbf{61.7} & 17.8 & \textbf{45.4} & \textbf{20.2} & \textbf{37.0} & \textbf{32.3} \\
\bottomrule
\end{tabular}}
\end{table}

The ablations show that language information is useful only when it is introduced with semantic structure and objectness coupling.
Label-only Prompt is unstable: it slightly improves Task 1 mAP from 59.5 to 59.7 and Task 2 Both AP from 44.0 to 44.1, but lowers several U-Recall and later-task aggregate entries.
This indicates that plain class names are weak anchors for OWOD because they lack appearance, context, and function cues for separating objects from background and unknown categories.
SETP is more reliable: extended descriptions raise Task 1 mAP to 60.7 and Task 2 Both AP to 44.8, showing that richer text prototypes help calibrate known-class logits.
However, SETP does not explicitly alter objectness, so its U-Recall gain remains moderate.
SPOF complements SETP by injecting known-semantic rejection into probabilistic objectness; it reaches the best Task 2 U-Recall of 18.0 and improves later-task Both AP over PROB.
The full \method{} model obtains the best overall trade-off across the reported columns, confirming that semantic alignment and semantic-probabilistic objectness fusion are complementary.

\subsection{Qualitative and Mechanistic Analysis}
The semantic branch gives an interpretable query-level score for known semantic support and known-semantic rejection.
We inspect high-objectness queries with low maximum similarity to known Extended-Text Embeddings and compare their predictions with PROB.
The desired behavior is that background regions remain filtered by probabilistic objectness, known objects retain high semantic support, and object-like regions unsupported by the known semantic space are preserved as unknown.
This analysis directly tests whether \method{} improves the OWOD decision boundary rather than merely increasing classification confidence.

\begin{figure}[t]
\centering
\includegraphics[width=0.52\linewidth]{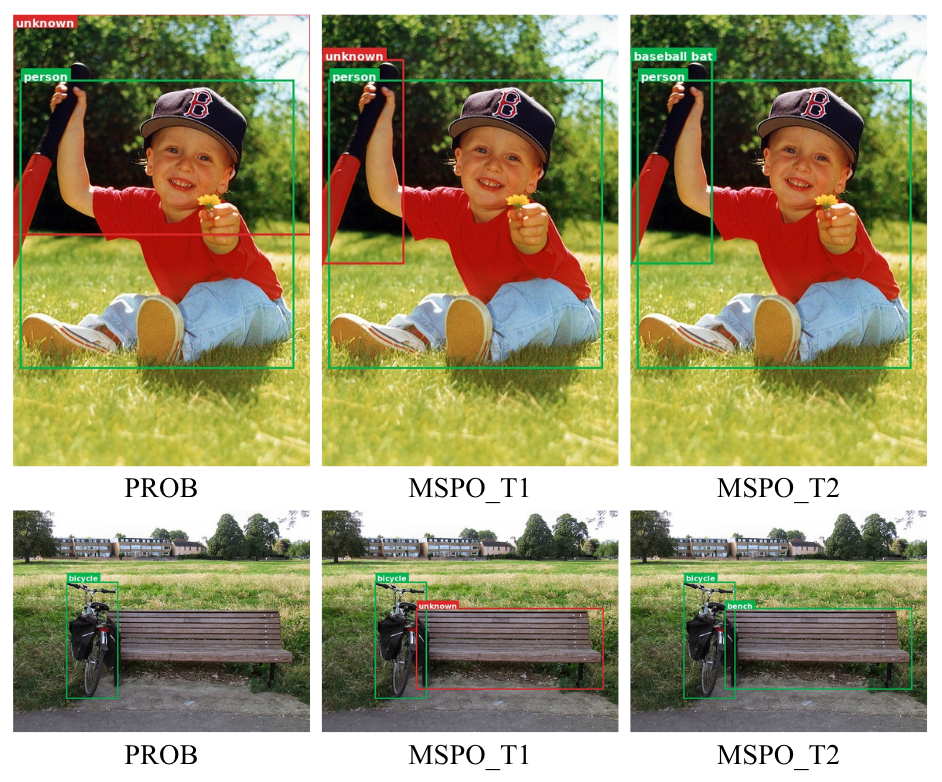}
\caption{Qualitative comparison between PROB and \method{} across incremental stages. In the first row, PROB in T1 incorrectly fires on background tree regions as unknown, whereas \method{} in T1 preserves the object-like baseball bat as unknown and recognizes it as baseball bat after the class is introduced in T2. In the second row, PROB detects the known bicycle but misses the bench as an unknown object; \method{} localizes the bench as unknown in T1 and recognizes it as bench after incremental training in T2.}
\label{fig:qualitative_cases}
\end{figure}

Figure~\ref{fig:qualitative_cases} illustrates how semantic calibration changes the unknown decision.
In the baseball-bat scene, PROB fires on background trees as unknown and misses the foreground bat, whereas \method{} suppresses the background response because low known-semantic support alone cannot overcome weak visual objectness.
The compact bat region is visually object-like but unsupported by current known semantics, so SPOF preserves it as unknown and later recognizes it after the class is introduced.
In the bench scene, PROB detects the known bicycle but misses the bench; \method{} instead treats the bench as an object-like region poorly explained by T1 known text prototypes.
These cases show that semantic-probabilistic objectness reduces background false unknowns while retaining semantically meaningful future objects.

\section{Conclusion}
\label{sec:conclusion}
We presented \method{}, a multimodal semantic extension of probabilistic objectness for open world object detection.
By constructing Extended-Text Embeddings, aligning decoder queries with the CLIP semantic space, estimating known-semantic support, and fusing this signal with probabilistic objectness, \method{} calibrates the boundary among known objects, unknown objects, and background while keeping the PROB detector intact.
Experiments and ablations show that known-category language semantics are a useful complement to probabilistic objectness under standard OWOD protocols.
These results suggest that OWOD objectness should not be treated as purely visual density estimation; calibrating object-like evidence by its support from the current known semantic space provides a lightweight and protocol-compatible improvement.

{\small
\bibliographystyle{splncs04}
\bibliography{main}
}

\end{document}